\documentclass[10pt, a4paper]{article}

\usepackage{lrec-coling2024} 

\usepackage{hyperref}
\usepackage{url}
\usepackage{times}
\usepackage{latexsym}
\usepackage{algorithm}
\usepackage{makecell}
\usepackage{CJKutf8}
\usepackage{color}
\usepackage{amssymb}
\usepackage{subfigure}
\usepackage{amsmath}
\usepackage{bbding}
\usepackage{algpseudocode}
\usepackage{algorithmicx,algorithm}

\usepackage{multirow}
\usepackage{booktabs}
\usepackage{arydshln}
\usepackage{graphicx}

\title{A Two-Stage Prediction-Aware Contrastive Learning Framework for Multi-Intent NLU}

\name{Guanhua Chen, Yutong Yao, Derek F. Wong$^\dagger$\thanks{$^\dagger$ Corresponding authors.}, Lidia S. Chao} 

\address{NLP\textsuperscript{2}CT Lab, Department of Computer and Information Science, University of Macau \\
         \{nlp2ct.guanhua, nlp2ct.yutong\}@gmail.com\\
         \{derekfw, lidiasc\}@um.edu.mo\\}

\abstract{
Multi-intent natural language understanding (NLU) presents a formidable challenge due to the model confusion arising from multiple intents within a single utterance. While previous works train the model contrastively to increase the margin between different multi-intent labels, they are less suited to the nuances of multi-intent NLU. They ignore the rich information between the shared intents, which is beneficial to constructing a better embedding space, especially in low-data scenarios. We introduce a two-stage Prediction-Aware Contrastive Learning (PACL) framework for multi-intent NLU to harness this valuable knowledge. Our approach capitalizes on shared intent information by integrating word-level pre-training and prediction-aware contrastive fine-tuning. We construct a pre-training dataset using a word-level data augmentation strategy. Subsequently, our framework dynamically assigns roles to instances during contrastive fine-tuning while introducing a prediction-aware contrastive loss to maximize the impact of contrastive learning. We present experimental results and empirical analysis conducted on three widely used datasets, demonstrating that our method surpasses the performance of three prominent baselines on both low-data and full-data scenarios.
 \\ \newline  }

\begin{document}

\maketitleabstract

\section{Introduction} \label{section.1}
Multi-intent Natural Language Understanding (NLU) models are fundamental building blocks within task-oriented dialogue systems \cite{qin-etal-2019-stack, gangadharaiah-narayanaswamy-2019-joint}. These systems encompass multi-intent detection (mID) and slot-filling (SF) tasks. However, effectively capturing multiple intents within short utterances presents a formidable challenge, primarily attributed to limited labeled data and the vast spectrum of spoken expressions. Consequently, a plethora of advanced models have emerged to refine the granularity of dialogue content and investigate the relationships among different intents \cite{qin-etal-2020-agif, song-etal-2022-enhancing}. Furthermore, recent researches \cite{qin-etal-2021-gl, 9747843, DBLP:conf/icassp/CaiZMF22, wu-etal-2022-incorporating} also proved that the extensive linguistic knowledge embedded within these pre-trained models, such as BERT \cite{devlin-etal-2019-bert} and RoBERTa \cite{DBLP:journals/corr/abs-1907-11692}, is effective to facilitate the comprehension of multiple intents within brief utterances.

However, these models often face a trade-off between inference speed and the incorporation of additional modules aimed at capturing more knowledge in the training data. This knowledge assumes a pivotal role in facilitating the model's acquisition of a more distinguishable semantic representation space, thereby yielding substantial benefits for downstream tasks. Consequently, several training strategies have emerged to maximize the utility of existing data without affecting the inference speed, such as contrastive learning \cite{liu-etal-2021-explicit-joint, basu-etal-2022-strategies} or curriculum learning \cite{10273589, zhou-etal-2020-uncertainty, zhan2021meta}. Leveraging these methods empowers the model to construct an improved embedding space while utilizing the same volume of data. \par

Regrettably, existing contrastive learning (CL) methods on multi-intent NLU typically assign fixed roles, either positive or negative, to samples. This potentially disregards the valuable knowledge embedded in the relationships between the shared intents, a factor that has been demonstrated to enhance multi-intent detection \cite{DBLP:conf/interspeech/XuS13}. Moreover, fine-tuning the model through equal-weight contrastive learning cannot fully learn the knowledge in the training data.

To tackle the above issues, we propose a two-stage Prediction-Aware Contrastive Learning (PACL) framework. PACL is meticulously designed to effectively harness knowledge emanating from shared intents through two-stage training: word-level pre-taining and prediction-aware contrastive fine-tuning. Building on \citet{DBLP:conf/interspeech/XuS13} proof of words commonly used to express an intent frequently appear across various instances, we further observed that these words exhibit a distinctive emphasis within the Part-of-Speech (POS) distribution. Hence, we introduce a word-level data augmentation strategy to construct a dataset for self-supervised pre-training. It enables the model to learn the associations between meaningful words and corresponding intents accurately. For fine-tuning stage, we devise an innovative prediction-aware contrastive learning framework. It facilitates automatic role-switching for each sample, allowing it to dynamically alternate between positive and negative roles while adjusting its influence based on the model's confidence. Additionally, we design an intent-slot attention mechanism to establish a strong connection between the mID and SF tasks for contrastive learning, which incentivized to more effectively harness the knowledge derived from shared intents, culminating in an embedding space characterized by heightened distinguishability.

We evaluate our PACL framework on three prominent multi-intent datasets: MixATIS \citelanguageresource{mixdata}, MixSNIPS \citelanguageresource{mixdata}, and StanfordLU \citelanguageresource{hou2020fewshot}. In addition, we employ three robust baselines for comparison: RoBERTa \cite{DBLP:journals/corr/abs-1907-11692}, TFMN \cite{chen-etal-2022-transformer}, and SLIM \cite{DBLP:conf/icassp/CaiZMF22}. Our experimental results demonstrate that PACL yields substantial enhancements in model performance while accelerating convergence speed. Our empirical analyses further reaffirm the indispensability of each component within our framework.

\section{Related work}
                                                                                                                                                                                                                                                                                                                                                                                                                                                                                                                                                                                                                                                                                                                                                                                                                                                                    \subsection{Multi-Intent NLU}

Compared to single-intent detection \cite{zhang-etal-2019-joint, wu-etal-2020-slotrefine, cheng2021effective}, multiple intents appear more common in real-time scenarios. Early research \cite{kim2017two,gangadharaiah-narayanaswamy-2019-joint} attempts to apply traditional CNN or RNN-based methods. \citet{qin-etal-2020-agif} proposed An Adaptive Graph-Interactive Framework (AGIF), and further expanded this technique to a non-autoregressive model \cite{qin-etal-2021-gl}. \citet{cai2022slim} proposed Explicit Slot-Intent Mapping with Bert (SLIM) which transferred JointBERT from single intent to multi-intent task while solving the shared-intent problem. Considering the number of intents is important in the multi-intent NLU task, \citet{chen-etal-2022-transformer} designed a novel threshold-free framework to predict the number of intentions in the utterance before predicting the specific intents.

\subsection{Contrastive Learning}
Contrastive Learning (CL) has been widely used on NLU tasks, because of its data scarcity and diverse expression. The recent works show the effect of CL on the NLU task \cite{gunel2020supervised, hou-etal-2021-learning, DBLP:conf/iclr/YehudaiVMLCC23}. Specifically, for the multi-intent NLU task, \citet{vulic-etal-2022-multi} devised a strategy to transform a general sentence-encoder into a task-specific one on multi-intent data through contrastive learning. \citet{DBLP:journals/corr/abs-2308-14654} proposed a novel bidirectional joint model trained using supervised CL and self-distillation, effectively utilizing intent and slot features to complement each other. 


\begin{figure}
    \centering
    \includegraphics[width=0.48\textwidth]{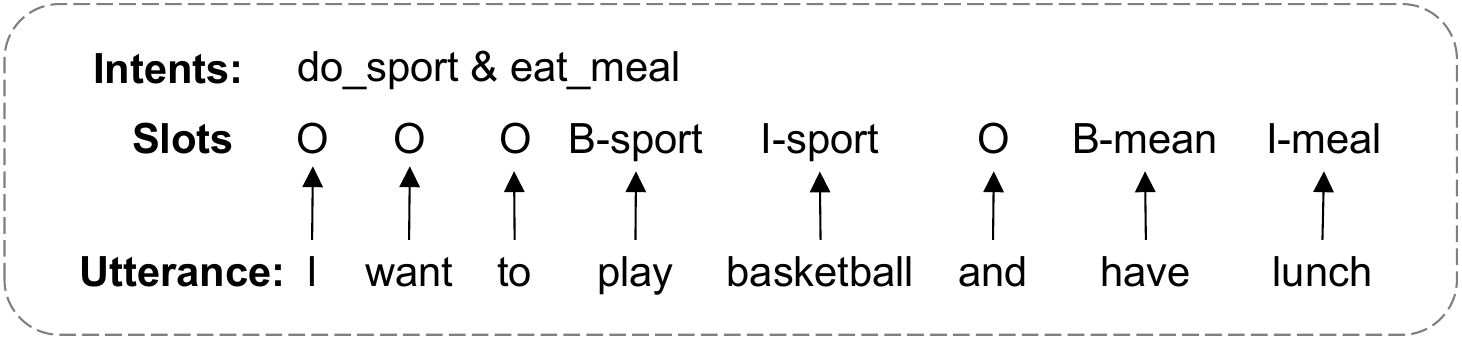}
    \caption{An example of multi-intent NLU task.}
    \label{fig.4}
\end{figure}

\section{Proposed Method}

\begin{figure*}
    \centering
    \includegraphics[width=0.98\textwidth]{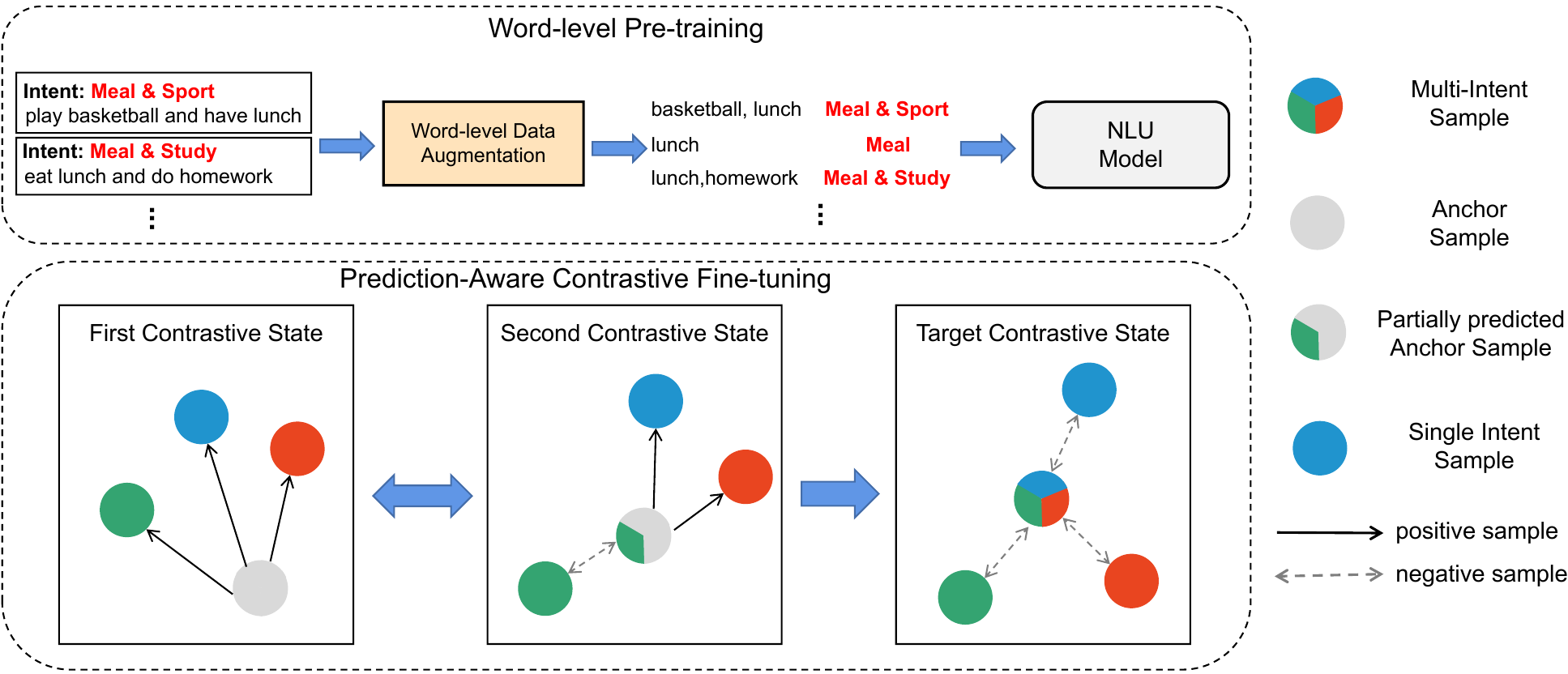}
    \caption{The overview of our framework. Different shapes indicate completely different samples and different shades of color indicate samples with shared intent.}
    \label{fig.1}
\end{figure*}

\subsection{Problem Formulation}
As illustrated in Figure \ref{fig.4}, given an input utterance $x$ = $(x_1, x_2, ..., x_n)$ with $n$ tokens, the multi-intent NLU model entails the simultaneous prediction of both multi-label intents for the utterance and the slot-filling for each word. Multi-label intent signifies the presence of more than one distinct intent within the set of possible intents.

The effectiveness of a joint training strategy for NLU task has been proved by \citet{DBLP:journals/corr/abs-1902-10909}. The joint objective functions can be mathematically formulated as follows:
\begin{equation} \label{eq.1}
    \mathcal{L}_{joint}= \mathcal{L}_{ID} + \mathcal{L}_{SF} 
\end{equation} 

\noindent where $\mathcal{L}_{ID}$ and $\mathcal{L}_{SF}$ are the cross entropy loss of intent detection and slot filling tasks. 

Building upon this foundation, we introduce our two-stage prediction-aware contrastive learning (PACL) framework as shown in Figure \ref{fig.1}.

\subsection{Word-level Pre-training}
As multi-intent samples tend to be particularly challenging to distinguish within the embedding space, our initial step involves word-level pre-training to bolster the model's adaptability to the specific domain. Based on \citet{DBLP:conf/icassp/CaiZMF22}, which released the correspondence between each token and sub-intent in the MixATIS \citelanguageresource{mixdata} and MixSNIPS \citelanguageresource{mixdata} datasets, we conducted an analysis focusing on the distribution of Part of Speech (POS) categories associated with tokens linked to distinct intents. Notably, POS categories like ``NN'', ``NNS'', and ``JJ'' accounted for a significant proportion of words connected to intent labels, reaching as high as 87.76\% in MixATIS and 78.97\% in MixSNIPS. This unveiled a strong correlation between specific POS categories and intents. We found that the intents are strongly correlated with some words with specific POS. \par

Thus, we split the original utterance-level dataset into word-level, concentrating on those aligned with the specified POS categories to construct a word-level multi-intent dataset. Firstly, we detect whether a word with the specific POS recurs across multiple utterances. Subsequently, we refined the intent associated with this word by extracting the shared intent from those recurring utterances. For instance, if a term ``lunch'' occurs in utterances linked to both ``$Meal \; \&\; Sport$'' and ``$Meal\; \& \; Study$'' intents, we identify it as belonging to ``$Meal$''. For words that remained connected to multiple intents, we concatenate them with words specifically indicating the associated intent. Since this is word-level pre-training, it is acceptable to have two words associated with the same intent in an input even after concatenation. 

Due to the absence of sentence structure information in the concatenation of words, our pre-training strategy exclusively focuses on the intent detection task. This phase is dedicated to enabling the model to acquire an understanding of the relationships between individual words and various intents, and facilitates the model's ability to more effectively capture the presence of multiple intents within an utterance. The final pre-training loss can be formulated as:

\begin{equation}
    \mathcal{L}_{PT} = \lambda_1\mathcal{L}_{ID}+\lambda_2\mathcal{L}_{CL}
\end{equation}

\noindent where $\mathcal{L}_{ID}$ is the cross-entropy loss of intent prediction and $\mathcal{L}_{CL}$ is the traditional contrastive loss of the intent logits. $\{\lambda_1, \lambda_2\}$ are the hyper-parameters to balance each loss.

\subsection{Prediction-Aware Contrastive Fine-tuning} \label{sec.2.1}
Upon the word-level pre-trained model, we introduce an innovative prediction-aware contrastive loss to fine-tune the model on the original utterance-level dataset. In this fine-tuning process, each instance, which shares common intents with other samples, dynamically alternates its role (either positive or negative). \par

As depicted in Figure \ref{fig.1}, given an anchor sample, samples with incorrectly predicted intents are designated as positive samples, while those with correctly predicted intents serve as negative ones. To illustrate, consider a sample with the intent ``$atis\_capacity\#atis\_city$'', once the model successfully predicts part of the intent, like ``$atis\_capacity$'', our model categorizes samples with ``$atis\_capacity\#atis\_city$'' and ``$atis\_capacity$'' as the positive candidates. Simultaneously, any samples with different intents, including ``$atis\_capacity$'',  are treated as negative. Because the anchor sample has already learned the knowledge of the relationship between the anchor sample and intent ``$atis\_capacity$''. \par

Noteworthy, those instance pairs with identical or completely different multi-intent labels will be regarded as mutually positive or negative pairs during the entire training. This prediction-aware contrastive learning empowers the model to gain a substantial understanding of the relationships between shared intents while expediting the clustering of samples sharing common intents.  \par

In greater detail, each time we construct the mini-batch, we sample a maximum of $K$ positive samples for each anchor sample. Regarding the negative samples, we randomly sampled them from the mini-batch. In situations involving a multi-intent label to which only the anchor sample belongs, we forward propagate this sample twice, following SimCSE \cite{gao-etal-2021-simcse}, to secure at least one positive sample for each anchor. The original CL objective function can be defined as:

\begin{equation} \label{eq.4}
        \mathcal{L}^i_{CL} =  \sum \limits_{p\in P(i)}\text{log}\frac{f(h_i, h_{i,p})}{\sum \limits_{k\in H(i)}f(h_i, h_{i,k})} 
\end{equation}

\noindent where $f(h_1,h_2)=\text{exp}(cos(h_1, h_2)/\tau)$. $P(i)$ is the set of positive samples, while $H(i)$ denotes the set of all positive and negative samples for the $i$-th anchor sample. $h_i$ is the representation vector for contrastive learning. The $\tau$ is the temperature value in CL loss. In our experiments, we sample at most $K$ positive or negative samples.

Nevertheless, contend that positive and negative samples with varying predicted probabilities should exert distinct influences on contrastive learning. We leverage the predicted probabilities to calibrate the impact of each sample within our prediction-aware contrastive loss function. In practice, the higher the average probability of the intents associated with the anchor sample, which matches that of the negative sample, the more challenging it becomes to distinguish the negative sample within the embedding space. Hence, we normalize the probability as the weight of negative samples as represented by the following equation:

\begin{equation} \label{eq.5}
    w_{i,n} = \text{Softmax}(Mean(\{p_{i}|y_{i}^{neg}\}))
\end{equation}
\noindent where $p_{i,j}$ is the probability that the $i$-th sample in the batch belongs to the $j$-th intent.  \par

With respect to the positive samples, a higher probability of a shared intent between the positive sample and the anchor sample signifies the model's enhanced capacity for precise classification. Consequently, it implies that the model can afford to allocate reduced attention to these samples. Correspondingly, we assign weights to the positive samples as follows: 

    \begin{equation} \label{eq.6}
\begin{split}  
    w_{i,p} & = \text{Softmax}(\alpha_i\prod \{p_{i}|y_{i}^{pos}\}), \\
    & \alpha_i = \frac{\text{Count}(y_{i}^{pos})}{\text{Count}(y_{i})}
\end{split}
\end{equation}

\noindent where the $\text{Count}(\cdot)$ means the number of elements that satisfy the condition, while $y_{i}$ refers to the predicted intent, and $y_{i}^{pos}$ represents the correctly predicted intent. Combining the equation \ref{eq.5} and \ref{eq.6}, the prediction-aware contrastive loss can be re-written as:

\begin{equation} \label{eq.7}
    \mathcal{L}^i_{PACL} =  \sum \limits_{p\in Pos(i)}\text{log}\frac{w_{i,p}f(h_i, h_{i,p})}{\sum \limits_{k\in H(i)}w_{i,k}f(h_i, h_{i,k})}
\end{equation}

Finally, in order to enhance the connection between multi-intent detection and slot filling tasks, we add a multi-head attention layer to capture the relationship between all the slots and intent representation. The final representation for CL can be generated as follows: 

\begin{equation}  \label{eq.2}
\begin{split}
    \text{Attention}(Q,K,V) = \text{Softmax}(\frac{QK^T}{\sqrt{d}})V, \\
        h_{intent} = \text{Attention}(h_{cls}, H_{slot}, H_{slot})
    \end{split}
\end{equation}
\begin{equation} \label{eq.3}
    h_{intent}' = (W[h_{intent}, h_{cls}]+b)
\end{equation}

\noindent where $\text{Attention}(\cdot)$ is the attention mechanism as described by \citet{DBLP:conf/nips/VaswaniSPUJGKP17}. $H_{slot}  \in \mathbb{R}^{n\times d}$ is the set of all the word representations of the given utterance. $W$ is the trainable parameters, and $h_{intent}' \in \mathbb{R}^d$ is the final representation. We feed the $h_{intent}'$ to the same intent classifier in equation \ref{eq.1} to train the additional layer in equation \ref{eq.2} and \ref{eq.3} only, which has few parameters and will not affect the training speed too much. This allows the latent space used for the CL to be more adapted to the intent classifier in equation \ref{eq.1}.

Overall, the final loss of the fine-tuning process can be formulated as:
\begin{equation}
    \mathcal{L}_{FT} = \lambda_3\mathcal{L}_{ID} + \lambda_4\mathcal{L}_{SF} + \lambda_5\mathcal{L}_{PACL}
\end{equation}
\noindent where $\{\lambda_3, \lambda_4, \lambda_5\}$ are the hyper-parameters to balance the impact of each loss.

\begin{table*}[t]
  \centering
    \renewcommand\arraystretch{1.05}
    \scalebox{1}{
    \begin{tabular}{l|c|c|c|c|c|c}
    \hline
     \multicolumn{1}{c|}{\multirow{2}{*}{Model}} & \multicolumn{3}{c|}{MixATIS} & \multicolumn{3}{c}{MixSNIPS}  \\
     \cline{2-7}
      \multicolumn{1}{c|}{}& IC Acc & SF F1 & Overall Acc & IC Acc & SF F1 & Overall Acc \\    
      \hline
      AGIF\cite{qin-etal-2020-agif} & 75.8 & 88.1 & 44.5 & 96.5 & 94.5 & 76.4 \\
      GL-GIN \cite{qin-etal-2021-gl} & 76.3 & 88.3 & 43.5 &95.6 & 94.9 & 75.4 \\
       GISCo \cite{song-etal-2022-enhancing} & - & - & 48.2 & - & - & 75.9 \\
      SDJN+BERT \cite{9747843} & 78.0 & 87.5 & 46.3 & 96.7 & 95.4 & 79.3 \\
      \hline
      \multicolumn{7}{c}{Low-Data} \\
      \hline
      RoBERTa$^\dagger$ \cite{DBLP:journals/corr/abs-1907-11692} & 46.7 & \textbf{84.1} & 24.9 & 94.0 & 92.3 & 67.5 \\
      RoBERTa (PACL) & \textbf{66.2} & 84.0 & \textbf{36.8}  & \textbf{94.8} & \textbf{92.9} & \textbf{69.6} \\
      \hline
       TFMN$^\dagger$ \cite{chen-etal-2022-transformer} & 46.4 & 71.1  & 15.7  & 95.2 & 90.9 & 65.2 \\
      TFMN (PACL)  & \textbf{58.3} & \textbf{72.7}  & \textbf{18.1}  & \textbf{95.5} & \textbf{91.9} & \textbf{67.8} \\
      \hline
       SLIM$^\dagger$ \cite{DBLP:conf/icassp/CaiZMF22}& 52.8 & 72.4  & 16.1  & 93.9 & 93.4  & 73.8 \\
      SLIM (PACL) & \textbf{60.8} & \textbf{75.8}  & \textbf{23.3}  & \textbf{95.2} & \textbf{94.4}  & \textbf{75.4} \\
    \hline
    \multicolumn{7}{c}{Full-Data} \\
      \hline
      RoBERTa$^\dagger$ \cite{DBLP:journals/corr/abs-1907-11692} & 77.5 & 85.5 & 47.7 & \textbf{96.5} & 95.6& 80.9 \\
      RoBERTa (PACL) & \textbf{79.1} & \textbf{86.0} & \textbf{48.9} & \textbf{96.5} & \textbf{96.2} &  \textbf{83.4} \\
      \hline
       TFMN$^\dagger$ \cite{chen-etal-2022-transformer} & 81.5 &  \textbf{86.9}  & 48.6 & 97.1 & 96.1 & 83.4 \\
      TFMN (PACL) & \underline{\textbf{82.9}} &  86.7  & \textbf{49.4} &  \underline{\textbf{97.4}} & \textbf{96.3} & \textbf{83.6} \\
      \hline
       SLIM$^\dagger$ \cite{DBLP:conf/icassp/CaiZMF22} & 78.9 & 87.1 & 46.4 & 96.8 & 96.3 & 83.7 \\
      SLIM (PACL) & \textbf{81.9} & \underline{\textbf{87.3}} & \underline{\textbf{50.4}} & \textbf{96.9} & \underline{\textbf{96.8}} & \underline{\textbf{85.1}} \\
    \hline
    \end{tabular}}
    \caption{The results (\%) on the test set of two datasets. $^\dagger$ means we reproduce this framework as our baselines. \textbf{Bold} numbers indicate the better result for each baseline, meanwhile, \underline{underlined} numbers show the best performance in the column.}
  \label{table1}
\end{table*}

\section{Experiments}

\subsection{Dataset and Evaluation}
We evaluate our method on three widely used multi-intent datasets, \textbf{MixATIS} \citelanguageresource{mixdata}, \textbf{MixSNIPS} \citelanguageresource{mixdata}, and \textbf{StanfordLU} \citelanguageresource{hou2020fewshot}. MixATIS contains 13161/756/828 utterances in train/validation/test set. MixSNIPS contains 39776/2198/2199 utterances in train/validation/test set. As for StanfordLU, it consists of 6428/790/820 train/validation/test data. We sampled 10\% of each intent for our low-data scenarios. For evaluation metrics, we use F1 score, intent accuracy, and intent-slot overall accuracy. 

\subsection{Baselines}
We compare our proposed approach with previous state-of-the-art models \cite{DBLP:conf/acl/ENCS19, qin-etal-2019-stack, qin-etal-2020-agif, qin-etal-2021-gl}, whose results are taken from previous work directly. Besides, we also reproduce three strong baselines: \textbf{RoBERTa}\footnote{https://huggingface.co/roberta-base} \cite{DBLP:journals/corr/abs-1907-11692}, \textbf{TFMN}\cite{chen-etal-2022-transformer}, and \textbf{SLIM}\footnote{https://github.com/TRUMANCFY/SLIM} \cite{DBLP:conf/icassp/CaiZMF22}, to verify the effect of our framework. More details about our implementations will be explained in Appendix.

\subsection{Implementation Detail} \label{appendix.1}
We use the BERT-Base\footnote{https://huggingface.co/bert-base-uncased} model for SLIM \cite{DBLP:conf/icassp/CaiZMF22} and TFMN \cite{chen-etal-2022-transformer}. Besides, we also employ the RoBERTa-Base\footnote{https://huggingface.co/roberta-base} \cite{DBLP:journals/corr/abs-1907-11692} model with 2 feed-forward classifier released by hugging face. The max sequence length and batch size are 50 and 64 for both pre-training and fine-tuning. The sampling number $K$ in our experiments is set to 5. We pre-train the model for $\{1, 3, 5\}$ epochs and fine-tune it for 10 epochs. For Roberta and TFMN models, we set the dropout rate and learning rate to 0.1 and 2e-5. For the SLIM model, we set the dropout rate and learning rate to 0.2 and 5e-5. For more details about training, given an input utterance with $n$ tokens, $x=\{x_1, x_2, ..., x_n\}$, and its encoded representation, $h=\{h_{cls}, h_1, h_2, ..., h_n, h_{sep}\}$, where the representation of the [CLS] token is utilized for multi-intent detection, and the representations of the other tokens are utilized for slot filling, separately.
\begin{table}[h]
  \centering
    \renewcommand\arraystretch{1.1}
    \scalebox{0.78}{
    \begin{tabular}{l|c|c|c}

    \hline
     \multicolumn{1}{c|}{\multirow{2}{*}{Model}} & \multicolumn{3}{c}{StanfordLU}   \\
     \cline{2-4}
      \multicolumn{1}{c|}{}& IC Acc & SF F1 & Overall Acc \\    
      \hline
      \multicolumn{4}{c}{Low-Data} \\
      \hline
      RoBERTa$^\dagger$ \cite{DBLP:journals/corr/abs-1907-11692} &21.3& \textbf{29.1}& 21.2 \\
      RoBERTa (PACL) &\textbf{32.4}&  28.8&  \textbf{31.9}  \\
      \hline
       TFMN$^\dagger$ \cite{chen-etal-2022-transformer} &41.7 &30.0 &38.8  \\
      TFMN (PACL) &\textbf{46.7} &\textbf{31.8}  &\textbf{40.2}   \\
    \hline
    \multicolumn{4}{c}{Full-Data} \\
      \hline
      RoBERTa$^\dagger$ \cite{DBLP:journals/corr/abs-1907-11692} &88.0& 92.3& 82.9 \\
      RoBERTa (PACL) &\textbf{89.0}& \textbf{92.5}& \textbf{84.1} \\
      \hline
       TFMN$^\dagger$ \cite{chen-etal-2022-transformer} & 88.0 & \textbf{93.0} & 83.6 \\
      TFMN (PACL) & \textbf{89.1}  & 92.9 & \textbf{84.3} \\
    \hline
    \end{tabular}}
    \caption{The results (\%) on the test set of StanfordLU dataset. $^\dagger$ means we reproduce this framework as our baselines. \textbf{Bold} numbers indicate the better result for each baseline.}
  \label{table2}
\end{table}

\begin{table*}[t]
  \centering
    \renewcommand\arraystretch{1.1}
    \scalebox{1.2}{
    \begin{tabular}{c|c|c|c|c|c|c|c|c}
    \hline
     \multicolumn{3}{c|}{\multirow{2}{*}{Method}} & \multicolumn{6}{c}{MixATIS}   \\
     \cline{4-9}
      \multicolumn{3}{c|}{}& \multicolumn{3}{c|}{Low-data}  &  \multicolumn{3}{c}{Full-data}    \\
     \hline
       PT & RS & PA & IC Acc & SF F1 & Overall Acc  &  IC Acc & SF F1 & Overall Acc\\    
      \hline
       \multicolumn{3}{c|}{Baseline}  & 52.8 & 72.4 & 16.1 & 78.9 & 87.1 & 46.4\\
       \hline
       - &- &- & 49.6 & 71.6 & 16.1  & 79.8 & 86.9 & 48.1  \\
       \hline
       $\surd$ &- &-  & 57.4 & 73.3 & 19.3 & 79.6 & 86.1  & 47.8  \\
       \hline
        - &$\surd$ &-  & 48.3 & 77.0 & 19.1 & 80.3 & 86.9 & 47.5  \\
        \hline
     - & - &$\surd$ & 50.1 & 71.5 & 15.6 & 80.1 & \textbf{87.6} & 48.1 \\
     \hline
    $\surd$ & $\surd$ & - & 54.7 & 75.4  & 21.7 & 80.5 &  86.9 & 48.9 \\
    \hline
     $\surd$  & - & $\surd$ & 54.5  & 75.4 & 21.3 & 79.9  & 86.4 & 49.3 \\
           \hline
      - &  $\surd$ & $\surd$  & 58.8 & 71.2 & 17.6 &  81.6 & 87.5  & 50.1   \\
      \hline
       $\surd$  &  $\surd$ & $\surd$&\textbf{60.8} & \textbf{75.8} & \textbf{23.3} & \textbf{81.9} & 87.3  & \textbf{50.4}   \\
    \hline
    \end{tabular}}
    \caption{The ablation study results(\%). None of the ``PT'', ``RS'', or ``PA'' components are used meaning the traditional CL method. More explanations are in Section \ref{sec.5.2}.}
  \label{table4}
\end{table*}

\subsection{Main Results}
  The primary experimental findings are summarized in Table \ref{table1} and Table \ref{table2}. Our proposed framework consistently outperforms RoBERTa \cite{DBLP:journals/corr/abs-1907-11692}, TFMN \cite{chen-etal-2022-transformer}, and SLIM \cite{DBLP:conf/icassp/CaiZMF22} on both low-data and full-data scenarios. Notably, we conducted experiments solely with the RoBERTa and TFMN models on the StanfordLU dataset, as this dataset lacks matching data for intent and individual tokens. Intuitively, the improvements achieved by the PACL framework are more substantial in low-data scenarios compared to high-data scenarios, across all three evaluation metrics. For instance, on the MixATIS dataset, our PACL method demonstrates a remarkable 11.9\% increase in intent detection accuracy and a 7.2\% boost in overall accuracy in the low-data scenario, whereas it only achieves an improvement of up to 2.7\% in intent detection accuracy and 3.7\% in overall accuracy in the high-data scenario. \par

Furthermore, we anticipate a modest enhancement in slot-filling F1 scores, as the primary objective of combining intent and slot representations is to primarily enhance overall accuracy. Consequently, we observe instances where the overall accuracy improved, while the individual scores for intent detection and slot filling exhibited less significant improvements. It is pertinent to mention that the improvement in intent classification accuracy is relatively lower on the MixSNIPS dataset compared to the other two datasets. This divergence may be attributed to the fact that MixATIS and StanfordLU comprise a significantly larger number of multi-intent labels, with 17 intents and 24 intents, in contrast to MixSNIPS, which contains only 6 intents. This discrepancy increases the proportion of overlapping intents, which, in turn, results in our PACL framework performing more effectively on the MixATIS and StanfordLU datasets.

\section{Analysis} 
\subsection{Ablation Study} \label{sec.5.2}
 For a better explanation, we divide the fine-tuning stage into two components in this section: dynamic role selection for each sample and the integration of prediction-aware contrastive loss. We conducted ablation experiments in both low-data and full-data scenarios to assess the efficacy of each component, as summarized in Table \ref{table4}. In this context, ``PT'' refers to the word-level pre-training process, ``RS'' signifies the dynamic selection of roles (positive or negative) for each sample during training, and ``PA'' indicates the prediction-aware contrastive loss. Our findings unequivocally establish the necessity of each component. Specifically, ``RS'' and ``PA'' can both contribute to enhancing the performance across all three metrics, while combining ``PA'' and ``RS'' obtains further improvement on overall performance. This indicates that these two components can benefit each other. Regarding ``PT'', it facilitates the model in acquiring domain-specific precise knowledge at word-level, consequently bolstering the effectiveness of the subsequent prediction-aware contrastive fine-tuning. However, ``PT'' gets less improvement on full data than on low-data, due to the fact that the greater amount of data in the full-data scenario allows the model to learn the knowledge between shared intents better, which diminishes the effect of ``PT''.

\begin{table}[t]
  \centering
    \renewcommand\arraystretch{1.1}
    \scalebox{1.05}{
    \begin{tabular}{c|c|c|c}
    \hline
     \multicolumn{1}{c|}{\multirow{2}{*}{Percentage}}  & \multicolumn{3}{c}{MixATIS}   \\
     \cline{2-4}
      \multicolumn{1}{c|}{}  & IC Acc & SF F1 & Overall Acc \\    
      \hline
       \multirow{2}{*}{10\%}  & 52.8 & 72.4 & 16.1 \\
       \cdashline{2-4}
        & \textbf{60.8} & \textbf{75.8} &\textbf{23.3}  \\
        \hline
       \multirow{2}{*}{20\%} & 71.4 & 82.1 & 36.9  \\
       \cdashline{2-4}
        & \textbf{72.6} & \textbf{84.2} &\textbf{40.2}  \\
        \hline
        \multirow{2}{*}{30\%} & 74.8 & 84.7 & 42.3\\
        \cdashline{2-4}
        & \textbf{75.9} & \textbf{86.2} &\textbf{44.0}  \\
        \hline
        \multirow{2}{*}{40\%} & 75.7 & 85.9 & 43.3  \\
        \cdashline{2-4}
         & \textbf{77.9} & \textbf{86.9} &\textbf{45.7}  \\
        \hline
        \multirow{2}{*}{50\%}  & 75.8 & 86.1 & 44.6 \\
        \cdashline{2-4}
        & \textbf{78.0} & \textbf{86.9} &\textbf{46.6}  \\
    \hline
    \end{tabular}}
    \caption{The experimental results with different percentages of the MixATIS dataset. For each percentage, the upper results are the SLIM baseline, while the bottom ones are our PACL framework. \textbf{Bold} numbers indicate the better result for each percentage.}
  \label{table5}
\end{table}

\subsection{Different Proportions of Training Data}
To further explore the effectiveness of our method on different data volumes, we conducted experiments on MixATIS dataset with \{20\%,30\%,40\%,50\%\} data, and the results compared with baselines are shown in Table \ref{table5}. Our method has an average 3\% improvement in intent accuracy as well as an average 3.28\% improvement in overall accuracy for data on different proportions. Intuitively, the enhancement of our method slightly decreases as the proportion of data increases. It is because more training data will enable the model to recognize multiple intents more clearly, which in turn diminish the effect of contrastive learning. Notably, the model performs much worse optimizing 10\% of training data than on the other proportions, meanwhile, PACL obtains the maximum performance optimization (8\% improvement on Intent Accuracy and 7.2\% improvement on Overall Accuracy) on 10\% of the data. This also proves that our PACL framework can help the model construct better embedding spaces indirectly, especially for low-data scenarios. 

\begin{figure}
    \centering
    \includegraphics[width=0.48\textwidth]{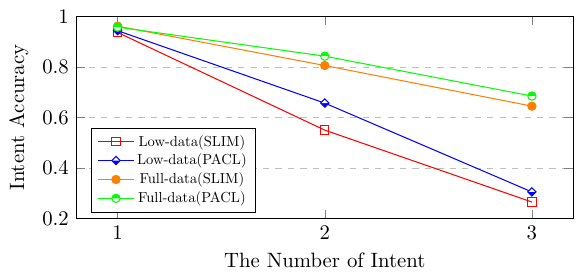}
    \caption{The Intent Accuracy of different numbers of intents trained on low- and high-data of MixATIS dataset.}
    \label{fig.5}
\end{figure}

\subsection{The Effect of Different Number Intents}
Next, we conduct a comparative analysis of the impact of the SLIM baseline and our PACL framework on labels with varying numbers of intents. As shown in Figure \ref{fig.5}, it is clear that on single-intent test data, there are minimal disparities in model performance across different methods and data volumes. However, the disparity between the performance of models trained on low- and high-data scenarios obtains significant improvements on test data with 2- and 3-intent labels. This divergence can be attributed to the fact that training on samples with multiple intents aids the model in comprehending individual intents, whereas the reverse is not as effective. Moreover, the degree of enhancement achieved by our method for the model intensifies with the increasing number of intents. In low-data scenarios, our approach enhances intent accuracy by 8\% for 2-intent test data and by 9\% for 3-intent data. In high-data scenarios, our method improves intent accuracy by 3.7\% for samples with 2-intent labels and by 4\% for samples with 3-intent labels. These results underscore the effectiveness of our PACL framework for multi-intent NLU while enhancing the efficiency of utilizing low-resource data.

\begin{figure}[t]
    \centering
    \includegraphics[width=0.48\textwidth]{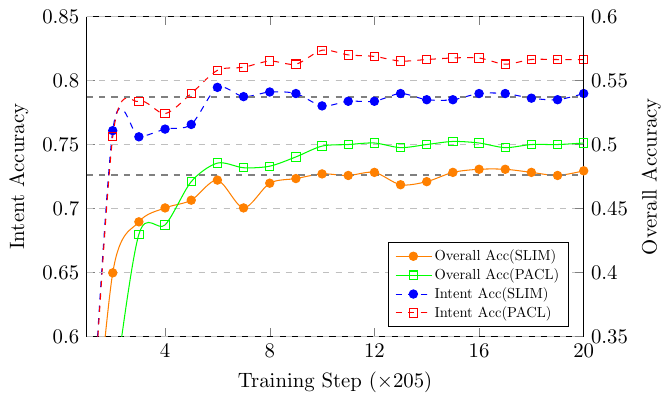}
    \caption{The learning curve of intent accuracy and overall accuracy on MixATIS test set every fixed validation step.}
    \label{fig.2}
\end{figure}

\subsection{Learning Curves}
In order to validate the impact of our PACL framework during the training process, we visually analyze the learning curve of SLIM \cite{DBLP:conf/icassp/CaiZMF22} on the test set of the MixATIS dataset (Figure \ref{fig.2}). The intent accuracy of the baseline and PACL approaches is denoted by dashed lines, while solid lines represent their respective overall accuracy curves. Our proposed method demonstrates swifter convergence compared to the baseline, showcasing consistent enhancements in both intent accuracy and overall accuracy. Notably, the overall accuracy of PACL in the early stage is lower than baseline. This makes sense because the pre-training process uses word-level inputs, and the model needs to adapt to the utterance-level pattern in the early stage of training process.

\begin{figure*}[t]
    \centering
    \includegraphics[width=0.48\textwidth]{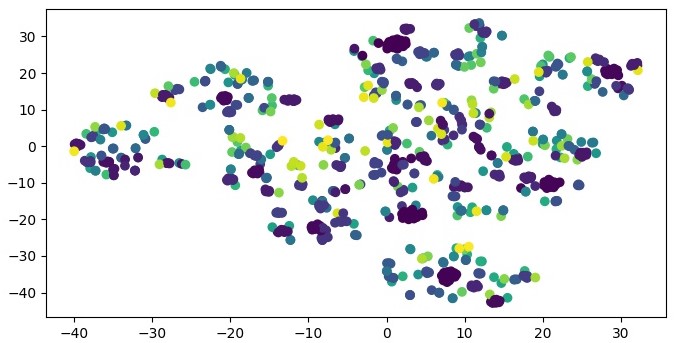}
    \includegraphics[width=0.48\textwidth]{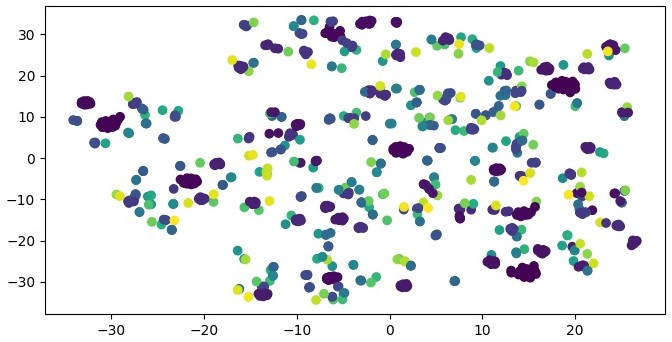}
    \caption{The distribution of intent embeddings in MixATIS dataset. The left one is trained on SLIM baseline, and the other one is trained with our PACL method.}
    \label{fig.3}
\end{figure*}

\begin{figure}[t]
    \centering
    \includegraphics[width=0.48\textwidth]{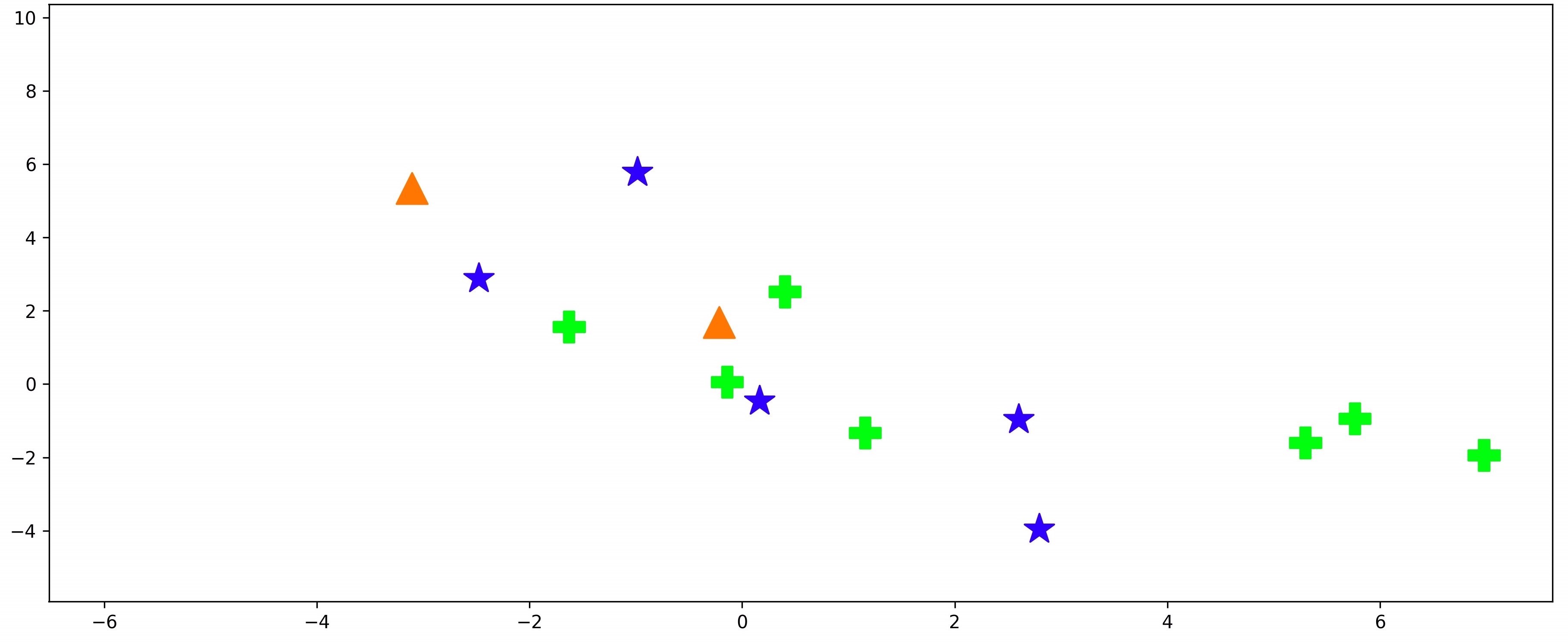}
    \includegraphics[width=0.48\textwidth]{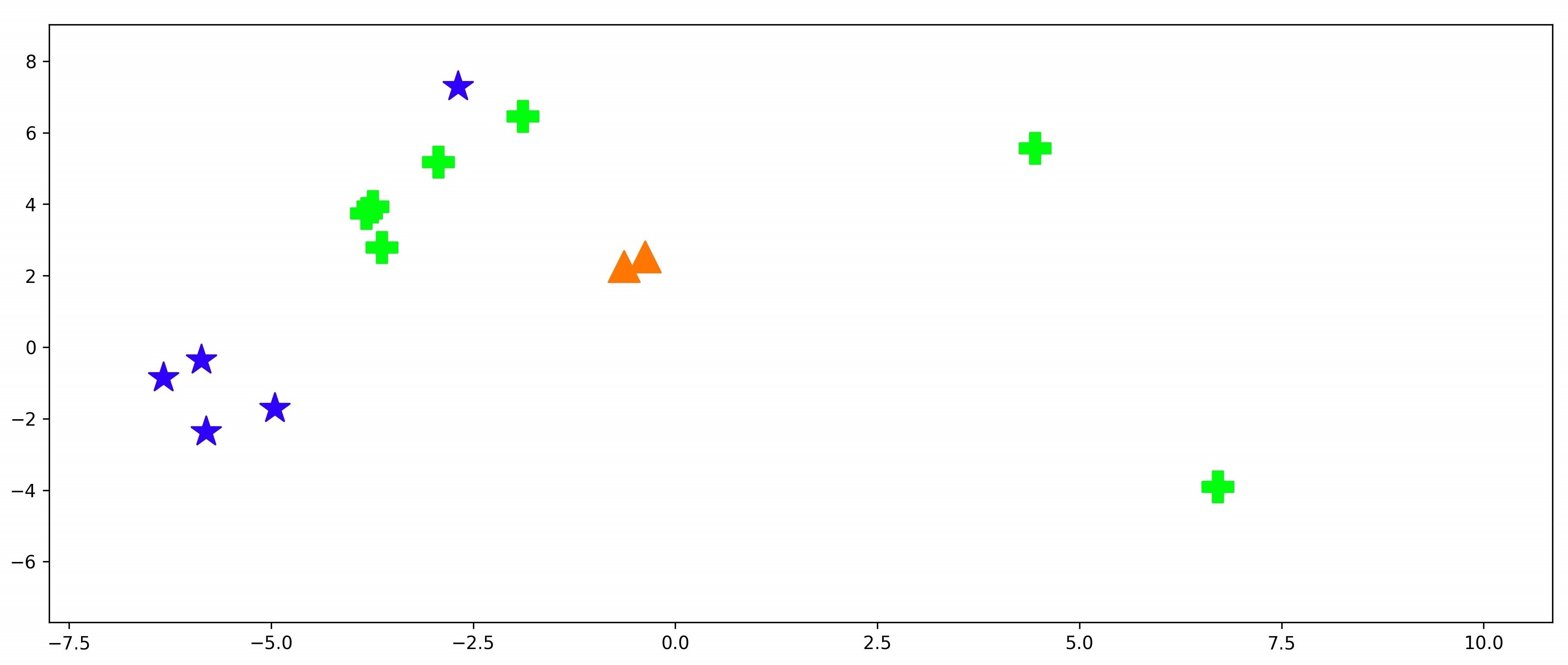}
    \caption{The distribution of several shared intents samples in MixATIS test set. The upper one is original BERT model, while the lower one is the BERT model pre-trained with our word-level pre-training strategy.}
    \label{fig.6}
\end{figure}

\subsection{Visualization}
As shown in Figure \ref{fig.3}, we utilized the t-SNE algorithm to visualize the distribution of intent embeddings in the MixATIS dataset. The embeddings derived from the SLIM baseline demonstrate the classification boundaries with limited capability to distinguish between shared intent samples. In contrast, the intent embeddings produced by our PACL model exhibit a more distinct distribution, indicating a heightened ability to differentiate between closely related intent classes. \par

In further detail, we randomly extract several samples with the same shared intents from the MixATIS test dataset to visualize the distinctions in embedding spaces between the original BERT model and the BERT model pre-trained using our word-level pre-training strategy. Notably, the visual analysis in Figure \ref{fig.6} reveals that, following our word-level pre-training strategy, the model exhibits the capacity to distinguish between utterances with distinct intents, even when they share some intents. This enhanced capability positions the model for improved performance during the fine-tuning stage.

\begin{table*}[t]
  \centering
    \renewcommand\arraystretch{1.05}
    \scalebox{0.82}{
    \begin{tabular}{l l}
    \hline
    \multicolumn{2}{c}{\textbf{Example 1}} \\
    \hline
    \textbf{Query}&  list la and also how many Canadian airlines flights use aircraft dh8 \\    
    \hline
    Model answer\\
    \hline
    \textbf{Intent}& atis\_city\#atis\_quantity\\
    \textbf{Slots}& O B-city\_name O O O O B-airline\_name I-airline\_name O O O B-aircraft\_code\\
    \hline
    Predicted by SLIM \cite{DBLP:conf/icassp/CaiZMF22}\\
    \hline
    \textbf{Intent}& \textcolor{red}{atis\_abbreviation}\#atis\_quantity\\
    \textbf{Slots}& O \textcolor{red}{B-aircraft\_code} O O O O B-airline\_name I-airline\_name O O O B-aircraft\_code\\
    \hline
    Predicted by PACL\\
    \hline
    \textbf{Intent}& \textcolor{blue}{atis\_city}\#atis\_quantity\\
    \textbf{Slots}& O \textcolor{blue}{B-city\_name} O O O O B-airline\_name I-airline\_name O O O B-aircraft\_code\\
    \hline
    \multicolumn{2}{c}{\textbf{Example 2}} \\
    \hline
    \textbf{Query}& \makecell[l]{book a restaurant for one person at 7 am and then play the album journeyman } \\  
    \hline
    Model answer\\
    \hline
    \textbf{Intent}& BookRestaurant\#SearchCreativeWork\\
    \textbf{Slots}& \makecell[l]{O O B-restaurant\_type O B-party\_size\_number O O B-timeRange I-timeRange O \\O O O B-object\_type B-object\_name}\\
    \hline
    Predicted by SLIM \cite{DBLP:conf/icassp/CaiZMF22}\\
    \hline
    \textbf{Intent}& BookRestaurant\#\textcolor{red}{PlayMusic}\#SearchCreativeWork\\
    \textbf{Slots}& \makecell[l]{O O B-restaurant\_type O B-party\_size\_number O O B-timeRange I-timeRange O \\O O O \textcolor{red}{B-music\_item B-album}}\\
    \hline
    Predicted by PACL\\
    \hline
    \textbf{Intent}& BookRestaurant\#SearchCreativeWork\\
    \textbf{Slots}& \makecell[l]{O O B-restaurant\_type O B-party\_size\_number O O B-timeRange I-timeRange O \\O O O \textcolor{blue}{B-object\_type B-object\_name}}\\
    \hline
    \end{tabular}}
    \caption{The examples of how our method solves the intent misclassification problem and improves overall accuracy. \textcolor{red}{Red} texts indicates the incorrect results, while \textcolor{blue}{blue} texts indicates the correct results.}
  \label{table3}
\end{table*}

\subsection{Case Study} \label{app.2}
 In addition to the t-SNE visualization, we exhibit two illustrative examples to validate the performance of our method more specifically, in which example 1 is drawn from the MixATIS dataset, while example 2 is drawn from the MixSNIPS dataset. As depicted in Table \ref{table3}, the SLIM baseline encounters challenges in making accurate predictions on both multi-intent detection and slot filling tasks, which primarily stems from the high similarity between multiple intents. In contrast, our method successfully demarcates a more distinct boundary, which ameliorates the mispredictions (Example 1) and underpredictions (Example 2) problems. Furthermore, our method addresses inaccuracies in slot predictions. The consistency of slot and intent corrections also demonstrates the strong correlation between the token and its corresponding intent.

\section{Conclusion}
 This paper introduces a novel two-stage prediction-aware contrastive learning framework for multi-intent NLU, which significantly enhances model performance through leveraging word-level pre-training and prediction-aware contrastive fine-tuning. Our method acquires knowledge not only from distinct intents, but also from shared common intents. The experimental results show that our method significantly improved for both low-data and full-data scenarios. As for the future work, we plan to explore how to maximize the impact of contrastive learning with a reduced number of positive samples.

 \section{Limitations}
The main limitation of this approach is training efficiency. Although it will not increase the inference time, it has more training cost because of the gradient calculation for contrastive samples, especially since the number $K$ is strongly related to the model performance. Besides, even though our approach utilizes the relation between the two tasks for associative contrastive learning, only intent labels are used for supervision. It might be better to design a strategy to introduce slot-label supervision.

\section{Acknowledgements}
This work was supported in part by the Science and Technology Development Fund, Macau SAR (Grant Nos. FDCT/060/2022/AFJ, FDCT/0070/2022/AMJ), National Natural Science Foundation of China (Grant No. 62261160648), Ministry of Science and Technology of China (Grant No. 2022YFE0204900), the Multi-year Research Grant from the University of Macau (Grant No. MYRG-GRG2023-00006-FST-UMDF), and Tencent AI Lab Rhino-Bird Gift Fund (Grant No. EF2023-00151-FST). This work was performed in part at SICC which is supported by SKL-IOTSC, and HPCC supported by ICTO of the University of Macau. 


\section{Bibliographical References}\label{sec:reference}
\bibliographystyle{lrec-coling2024-natbib}
\bibliography{lrec-coling2024-example, anthology}

\section{Language Resource References}
\label{lr:ref}
\bibliographystylelanguageresource{lrec-coling2024-natbib}
\bibliographylanguageresource{languageresource}

\end{document}